\documentclass{article}

\PassOptionsToPackage{numbers, compress}{natbib}
\usepackage[preprint]{neurips_2025}

\usepackage[utf8]{inputenc} % allow utf-8 input
\usepackage[T1]{fontenc}    % use 8-bit T1 fonts
\usepackage{hyperref}       % hyperlinks
\usepackage{url}            % simple URL typesetting
\usepackage{booktabs}       % professional-quality tables
\usepackage{amsfonts}       % blackboard math symbols
\usepackage{nicefrac}       % compact symbols for 1/2, etc.
\usepackage{microtype}      % microtypography
\usepackage{lipsum}		% Can be removed after putting your text content
\usepackage{graphicx}
\usepackage{natbib}
\usepackage{doi}
\usepackage{algorithmicx,algorithm,algpseudocode}

\usepackage[utf8]{inputenc} % allow utf-8 input
\usepackage[T1]{fontenc}    % use 8-bit T1 fonts

\usepackage{xcolor}         % colors
\usepackage{amsmath, amssymb, amsthm}
\usepackage{subcaption}
\usepackage{graphicx}
\usepackage{todonotes}
\usepackage{multirow}
\usepackage{colortbl}
\usepackage{wrapfig}
\usepackage{afterpage}
\usepackage{pifont}

\usepackage{amsthm}

\makeatletter
\newcommand{\shortmethod}[1][]{%
  \mathcal{R}_{\mathrm{S}}%
  \ifx\relax#1\relax
  \else
    ^{(#1)}%
  \fi
}
\makeatother

\usepackage{listings}
\usepackage[dvipsnames]{xcolor}

\title{Learning effective pruning at initialization \\ from iterative pruning}

\author{%
    Shengkai Liu$^{1,2,}$\thanks{Joint first authorship}\; , Yaofeng Cheng$^{1,2,}$\footnotemark[1]\; , Fusheng Zha$^{1,2}$\;, Wei Guo $^{1,2}$\;, \\[0.2em] \textbf{Lining Sun}$^{1,2}$\;, \textbf{Zhenshan Bing}$^{3}$\;, \textbf{Chenguang Yang}$^{4}$\;,
    \\[0.2em]
    $^1$State Key Laboratory of Robotics and Systems \; $^2$Harbin Institute of Technology \; \\ $^3$Technical University of Munich\; $^4$University of Liverpool \;
    \\[0.4em]
    \href{https://github.com/ChengYaofeng/AutoSparse.git}{\texttt{https://github.com/ChengYaofeng/AutoSparse.git}} 
}

\begin{document}

\maketitle

\begin{abstract}
\label{abst}

Pruning at initialization (PaI) reduces training costs by removing weights before training, which becomes increasingly crucial with the growing network size. However, current PaI methods still have a large accuracy gap with iterative pruning, especially at high sparsity levels. This raises an intriguing question: can we get inspiration from iterative pruning to improve the PaI performance? In the \textit{lottery ticket hypothesis}, the iterative rewind pruning (IRP) finds subnetworks retroactively by rewinding the parameter to the original initialization in every pruning iteration, which means all the subnetworks are based on the initial state. Here, we hypothesise the surviving subnetworks are more important and bridge the initial feature and their surviving score as the PaI criterion. We employ an end-to-end neural network (\textbf{AutoS}parse) to learn this correlation, input the model's initial features, output their score and then prune the lowest score parameters before training. To validate the accuracy and generalization of our method, we performed PaI across various models. Results show that our approach outperforms existing methods in high-sparsity settings. Notably, as the underlying logic of model pruning is consistent in different models, only one-time IRP on one model is needed (e.g., once IRP on ResNet-18/CIFAR-10, AutoS can be generalized to VGG-16/CIFAR-10, ResNet-18/TinyImageNet, et al.). As the first neural network-based PaI method, we conduct extensive experiments to validate the factors influencing this approach. These results reveal the learning tendencies of neural networks and provide new insights into our understanding and research of PaI from a practical perspective. Our code is available at: \url{https://github.com/ChengYaofeng/AutoSparse.git}.

\end{abstract}

\section{Introduction}
\label{intro}

Neural network pruning, a technique employed for several decades \citep{lecun1989optimal, reed1993pruning, blalock2020state}, involves selectively removing non-essential parameters from a network. This process maintains the network's inference accuracy while reducing the computational demands. This is particularly beneficial in resource-constrained environments such as mobile devices as it allows for faster response times and reduces energy consumption. While model pruning is traditionally performed after training to improve inference speed without compromising accuracy \citep{gale2019state, zhu2017prune, han2015learning}, the growing number of parameters in neural network models has led to a significant increase in training resource consumption \citep{brown2020language}. This shift has spurred interest in strategies for pruning networks at the early stages of training.

Recently, the Lottery Ticket Hypothesis (LTH) \citep{frankle2018lottery} has revealed that it is possible to identify efficient subnetworks early in the training process. These subnetworks can achieve accuracy levels comparable to those achieved after full training. Identifying these subnetworks early on could drastically reduce training time and resource usage, providing an efficient pathway to model optimization. Therefore, some approaches \citep{lee2018snip, wang2020picking, tanaka2020pruning, hoang2023revisiting} explore pruning at initialization (PaI), direct remove weights before training by assessing parameter importance using features before training. However, a comparative study \citep{frankle2020pruning} has shown that these PaI methods, using handcrafted criteria to prune, often do not match the performance of traditional iterative pruning, particularly at high sparsity.

The iterative pruning process of LTH \citep{frankle2018lottery} involves one significant step: rewind, which reinits the pruning subnetwork to its original initialization and then starts the next pruning iteration. We refer to this pruning method as Iteration Rewind Pruning (IRP), which leads to two key insights: a) All parameters are changed back to their initial values before further training, indicating that the subnetwork is closely related to their initial state. b) Each pruning cycle aims to find the best subnetwork based on the surviving parameters, relatively less important parameters are progressively pruned, resulting in the retention of parameters that represent a higher level of importance.

This motivated us to use the parameter's surviving iteration as parameter scores and investigate the correlations between this and their initialized features. We experiment with the IRP using LeNet-300-100 on MNIST and analyse the results. The initialized parameters and their scores are visualized in Fig.1. It seems the magnitude of the parameters does not have an intuitive correlation with their importance. We further investigate this importance using other PaI criteria (e.g. SNIP --- connection sensitivity and GraSP --- gradient flow) in Fig.1. The results are inconsistent. The above raises intriguing research questions: Does this surviving score from iterative pruning make sense? Due to the unintuitive correlation, how to build the relation initial feature with this score?

In this paper, we propose \textbf{AutoS}parse, a data-driven framework to predict the above scores before training, which automatically learns PaI criteria from iterative pruning. The network takes initial features, such as initial parameters and initial gradients on the dataset as inputs, and outputs the scores of parameters. Parameters with the lowest scores are then pruned according to the desired sparsity level. Comprehensive experiments evaluate the effectiveness and generalization of our method. Surprisingly, we find the results outperform recent state-of-the-art methods. Notably, although we need iterative pruning to create a dataset, this only needs once. Similar to existing pruning methods \citep{lee2018snip, wang2020picking} one criterion can be applied to all models, and our experiments demonstrate the ont-time RIP from one model can teach the PaI criterion for all the models. As the first data-driven criterion for PaI, we conduct extensive experiments across various aspects (e.g., datasets, inputs, models) to explore its effectiveness. Different from previous theoretical methods, these findings reveal the learning tendencies of neural networks, which advance the understanding and further exploration of PaI from a practical perspective. Our contribution can be summarized as follows:

\begin{itemize}
	\item{We propose a novel PaI parameter importance criterion through iterative rewind pruning and investigate its characteristics, highlighting the complexity of designing such criteria manually.}
	
	\item{Based on the importance criterion, a novel PaI approach (AutoS) is proposed, utilizing an end-to-end neural network to determine which parameter to prune. This approach introduces a transformative way of thinking in the field by shifting from traditional human-designed criteria approaches to data-driven, automated pruning.}

	\item{Extensive experiments demonstrate that AutoS can achieve high PaI accuracy, outperforming other baselines. This also shows that data-driven PaI methods can achieve high performance.}
	
	\item{As the first data-driven PaI criterion, comprehensive ablations and analyses evaluate the factors influencing this approach and the results advance the understanding and exploration of PaI.}
\end{itemize}

\section{Related Work}
\label{sec:related}

Neural network pruning encompasses a variety of methodologies characterized by diverse approaches \citep{wimmer2023dimensionality} such as structured \citep{liu2018rethinking} or unstructured pruning \citep{li2017pruning, mao2017exploring}, global \citep{frankle2018lottery, wang2020picking} or layer-wise pruning \citep{mocanu2018scalable, evci2020rigging}, and differences in pruning frequency (pruning at initialization \citep{lee2018snip} or iterative pruning \citep{frankle2018lottery, tanaka2020pruning}). To enhance the clarity of this study, our analysis specifically concentrates on the following two parts.

\paragraph*{After training \& Before training}
Most existing pruning methods assign scores to parameters after training, removing those with the lowest scores \citep{dong2017learning, molchanov2016pruning}. This approach is effective as parameter values stabilize after training, simplifying the assessment of their importance. The criteria typically include parameter magnitudes \citep{han2015learning}, impact on loss \citep{lecun1989optimal}, and various complex coefficient \citep{wang2019eigendamage, yu2018nisp, guo2016dynamic}. However, these algorithms primarily enhance inference efficiency without reducing the computational demands during training. The LTH \citep{frankle2018lottery} demonstrates that certain subnetworks, identifiable before training, can match the performance of a dense model. Consequently, recent research has focused on developing criteria to effectively identify these subnetworks at initialization. This work \citep{liu2022unreasonable} evaluates that random pruning is effective when the network size and layer-wise sparsity ratios are appropriate. Several methods propose more universal criteria to predict parameter scores before training, such as connection sensitivity in SNIP \citep{lee2018snip}, gradient flow in GraSP \citep{wang2020picking}, and synaptic sensitivity in Synflow \citep{tanaka2020pruning}. Additionally, recent findings suggest that the Ramanujan Graph criterion is also beneficial for pruning before training \citep{hoang2023revisiting}. This work focuses on pruning before training.

\paragraph*{Pruning at Initialization \& Iterative Pruning}

The LTH \citep{frankle2018lottery} introduces a classical iterative pruning process that involves training, pruning, and rewinding. This cycle is repeated until an optimal subnetwork is identified, characterized by a mask that signifies an efficient subnetwork specific to the initialized model and dataset. Despite its high accuracy, iterative pruning is resource-intensive, prompting recent research toward less costly alternatives, such as PaI. Techniques such as SNIP \citep{lee2018snip} and GraSP \citep{wang2020picking} implement pruning in a single step at initialization. In contrast, some approaches attempt a more labour strategy by pruning iteratively before training, allowing for cumulative data analysis. SNIP-it \citep{verdenius2020pruning} iteratively tests using a small batch, utilizing feedback from this batch to refine the entire model. Synflow \citep{tanaka2020pruning} iterative pruning can prevent layer collapse and proposes data-agnostic criteria that facilitate rapid iterations in predicting parameter scores. While PaI is faster, there is still a performance gap compared to iterative pruning, especially at high levels of sparsity. Our motivation is to minimize the performance discrepancy between PaI and iterative pruning.

\section{Analysis of Surviving Iteration}

\subsection{Preliminary}

\paragraph{Pruning Formulation}

Suppose we have a neural network parameterized by $\theta \in \mathbb{R}^k$ with its corresponding pruning masks $m \in {\{ 0, 1 \}}^k$. A pruned subnetwork can be generated with $\theta \odot m$, where $\odot$ denotes the Hadamard product. Given a target dataset $\mathcal{D}=\left\{\left(\mathbf{x}_i, \mathbf{y}_i\right)\right\}_{i=1}^N$ and a desired sparsity level $\epsilon \in [0,1]$, the pruning problem can be formulated as:

\begin{equation}
\label{equa_1}
\min _{m, \theta}  L(m \odot \theta ; \mathcal{D}) = \min _{m, \theta} \frac{1}{n} \sum_{i=1}^n \ell\left(m \odot \theta ;\left(\mathbf{x}_i, \mathbf{y}_i\right)\right), \quad  \text { s.t. } \theta  \in \mathbb{R}^k,  \|\mathbf{m}\|_0 \leq \epsilon,
\end{equation}

where $\ell$ is the loss function. The formulation of PaI is to select a pruning mask before training. Therefore, the score $s$ of each parameter is used to define which parameter should be pruned based on the initialized feature as follows:

\begin{equation}
\label{equa_2}
s \leftarrow \left(\mathcal{F}_0; \mathcal{D} \right)
\end{equation}

Here, $\mathcal{F}_0$ is features obtained before training, e.g. the initialized parameter ${\theta}_0$ and the initial gradient on dataset $g_0(\mathcal{D})$.

\begin{figure*}[!htb]\centering
	\includegraphics[width=5.3in]{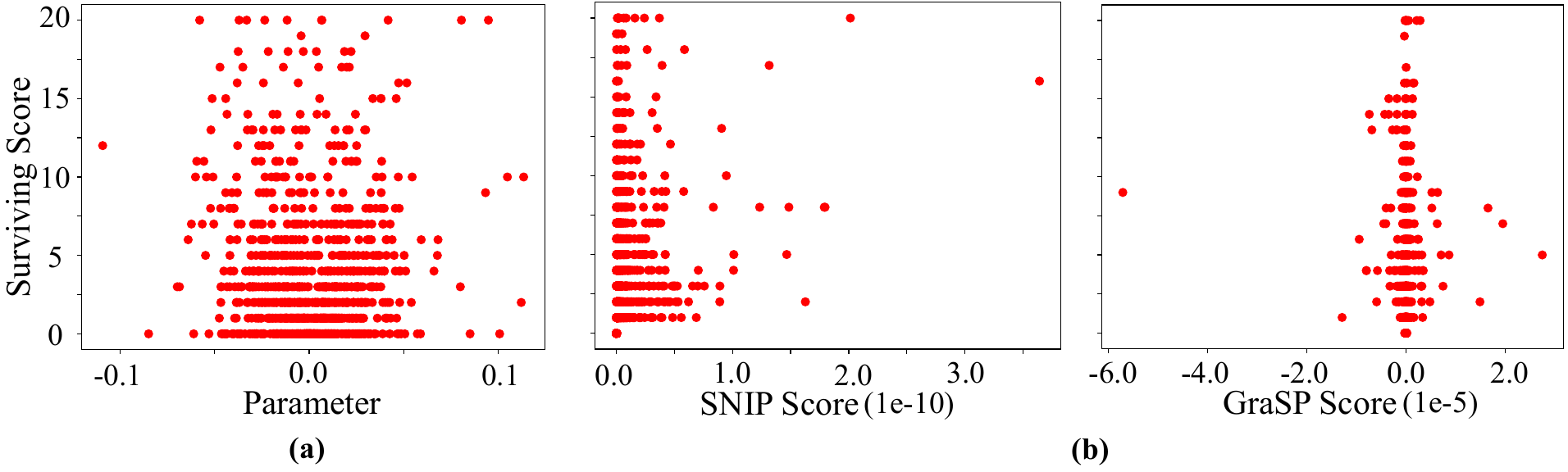}
	\caption{Visualization of Surviving Score. The score is obtained from IRP on LeNet-300-100 of MNIST. (a) The correlation between surviving score and initial parameter. (b) Comparing of surviving score and SNIP \citep{lee2018snip} and GraSP \citep{wang2020picking}. }
	\label{fig_1}
\end{figure*}

\subsection{Initial Features and Surviving Iteration}

\begin{figure*}[!htb]\centering
	\includegraphics[width=5.3in]{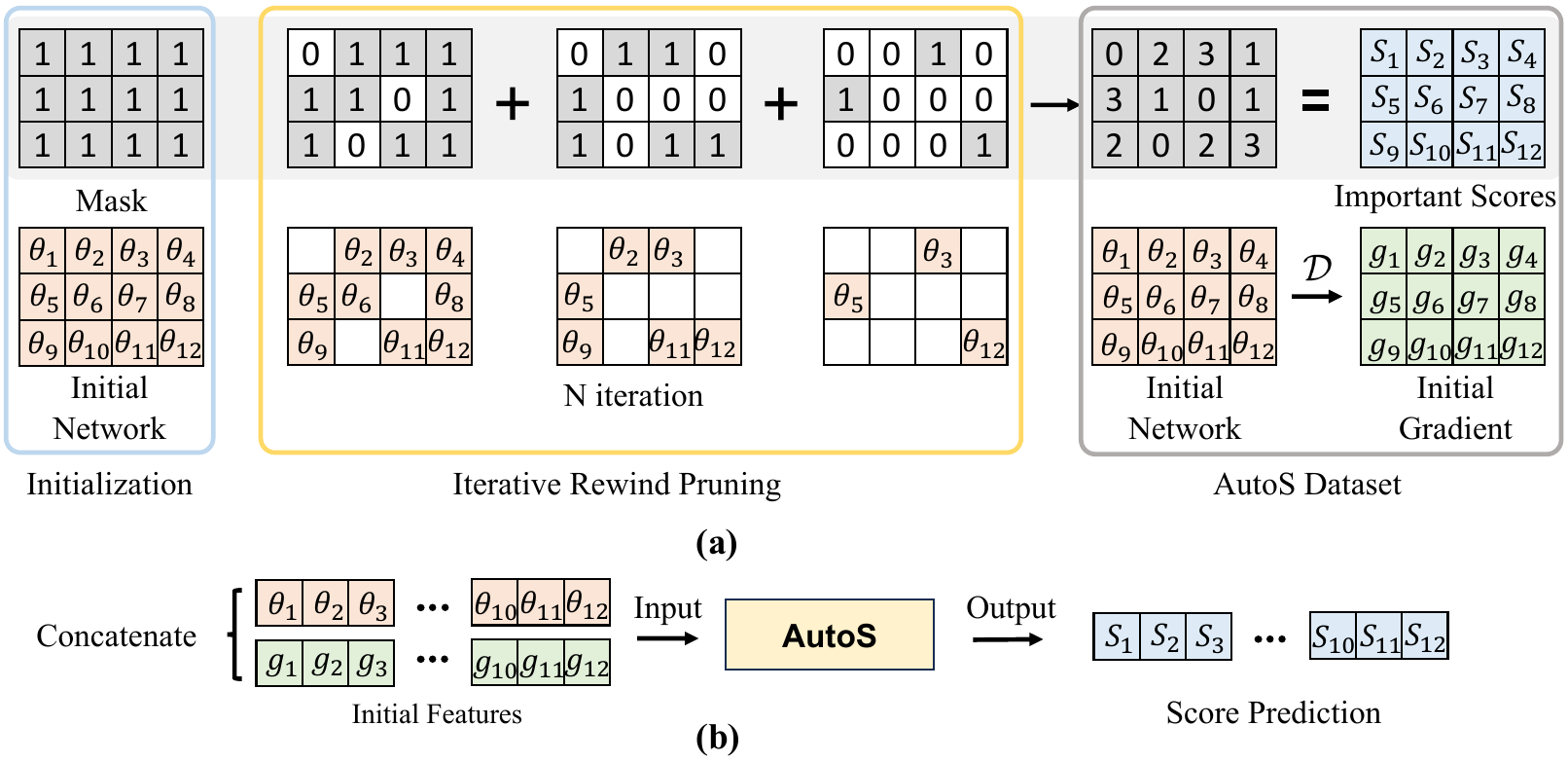}
	\caption{(a) illustrates the process of obtaining the surviving score, which also serves as the generation process for the AutoS dataset $\mathcal{D}_{\mathcal{A}}$. $\mathcal{D}_{\mathcal{A}}$ consists of parameters after initialization and gradients obtained based on the pruning dataset. (b) demonstrates how AutoS predicts these scores. Specifically, features from the initial network are concatenated and input into AutoS, which then predicts the parameter scores.}
	\label{fig_2}
\end{figure*}

Figure \ref{fig_2}(a) illustrates the process for acquiring the importance scores of the surviving subnetworks. We subsequently conducted several experiments to verify the correlation between these scores and the initial features. Since the IRP of LTH involves pruning based on the magnitude of the parameters, we first visualized the initial parameters and the corresponding scores in Figure \ref{fig_1}(a). The relationship, as observed, is not immediately intuitive. We then visualized the scores obtained from two other methods, SNIP and GraSP, in Figure \ref{fig_1}(b). The results do not show a positive correlation. However, these methods are still effective in achieving PaI. This raises the question: Is the surviving score truly meaningful?

Given the unintuitive correlation between the surviving score and the parameters, we avoided designing handcrafted criteria. Instead, we employed a neural network to learn the relationship between the initial features and the score. Surprisingly, the PaI accuracy at high levels of sparsity surpassed that of state-of-the-art methods. Details of the AutoS method can be found in Sec. \ref{method}. AutoS, an end-to-end network, uses the features before training to predict the score of each parameter, subsequently pruning the lowest-scoring parameters based on specified sparsity. This approach not only shifts theoretical exploration into practical application, making it easier to identify ways to improve PaI accuracy, but also provides more intuitive insights through visualization of the network's learning process. This visualization can offer valuable information on what the network focuses on, potentially inspiring further research.

\section{Methods}
\label{method}

\paragraph{Baseline PaI Method} We list some previous methods following as our baseline and to show our motivation. For details of the replication, see Appendix.

\textit{SNIP} \citep{lee2018snip}: this method computes the influence on loss as scores. It calculates the gradient in the dataset for every parameter as the score $S = |g (\theta; \mathcal{D})|$, and removes the parameters with the lowest score.

\textit{GraSP} \citep{wang2019eigendamage}: this method thinks the gradient flow is more important, and calculates the parameter score with Hessian-matrix $S = -\theta \odot \mathbf{H}g$.

\textit{NPB} \citep{pham2024towards}: this method thinks the effective paths and nodes will improve the pruning performance, and use $S = \alpha f_n+(1-\alpha) f_p$ to decide the value of a parameter, where $f_n$ is effective nodes and $f_p$ is effective paths of each parameter.

\textit{Random}: this method removes the parameter randomly after parameter initialization before training.

\subsection{Overview}

In this section, we will introduce how the dataset is created and the network of AutoS, as shown in Figure 2. Given initialized parameters ${\theta}_0$ and initial gradient on the dataset $g_0(\mathcal{D})$, an end-to-end network to predict the parameter's score. Once we have obtained the importance scores, we prune the parameters with the lowest scores to achieve the desired sparsity. Figure \ref{fig_2} (b) illustrates this end-to-end prediction process and pseudocode can be found in Algorithm \ref{code_pruning}. The predicted score can be written as:

\begin{equation}
\label{equa_3}
S_{surv} = \text{AutoS} \left(\theta_0, g_0(\mathcal{D}) \right) 
\end{equation}

\begin{algorithm}
\caption{AutoS Pruning Pipeline}
\label{code_pruning}
\begin{algorithmic}[1]
\Require  \\
Dataset $\mathcal{D}$: Pruning Dataset; \\
Network: a network parameterized by $\theta \in \mathbb{R}^k$; \\
Mask: corresponding pruning masks $m \in {\{ 0, 1 \}}^k$; \\
Target Sparsity $\epsilon$: desired sparsity level $\epsilon \in [0,1]$;\\
AutoS model: an \textit{AutoS} model for PaI;

\renewcommand{\algorithmicrequire}{\textbf{Input:}}
\Require $\theta_0$: network initialize; $m_0$: mask initialize
	\State \textit{model\_eval}: $g_0(\mathcal{D}) \leftarrow (\theta_0, \mathcal{D})$; \hfill model eval for initial gradient
	\State $S_{surv} = AutoS(\theta_0, g_0(\mathcal{D}))$ \hfill score prediction
	\State $m_{targ} \leftarrow (m_0, S_{surv}, \epsilon)$ \hfill pruning mask
	\State $\theta_{targ} \leftarrow (\theta_0, m_{targ})$ \hfill pruning

\renewcommand{\algorithmicrequire}{\textbf{Output:}}
\Require {Pruning Network $\theta_{targ}$}
\end{algorithmic}
\end{algorithm}

\subsection{Dataset Generation and Training}

To use AutoS for predicting the importance scores of a network, we need to acquire the ground truth dataset $\mathcal{D}_{\mathcal{A}}$. Figure \ref{fig_2} (a) illustrates this pipeline. Notably, the pruning rules of CNN and MLP are consistent to different models, the same as the related PaI methods \citep{lee2018snip, wang2020picking, pham2024towards} that one handcrafted rule can be applied to different models. Thus the datasets acquired from only one-time IRP can train AutoS that can be generalized to unknown models. In our approach, we employ the SNIP \citep{lee2018snip} to prune iteratively on ResNet-18 (CIFAR-10) once. Specifically, each parameter is assigned a binary mask $m \in \{0, 1\}^k$ initialized to $m_0 = 1$. We define the sparsity degree $\epsilon$, and the number of sparse iterations $N$ for the pruning process. This iterative process results in a mask $m_i$ after $i (i \leq N)$ times pruning. During each pruning iteration, certain parameters are removed, and their corresponding mask values are updated to 0. The importance of each parameter, denoted as $S_{surv}$, is set according to the number of training rounds it survives, which can be formulated as:

\begin{equation}
\label{equa_4}
S_{surv}^{gt} = \text{normlize}\sum^N_i m_i
\end{equation}

The pseudocode is shown as Algorithm \ref{code_datagen}. After finishing the dataset, we choose ResNet \citep{he2016deep} models as the AutoS backbone and add an encode layer to match the input features, a decode layer for the score prediction. For full model details, see Appendix \ref{autos_network}.

\section{Experiment}
\label{exp}

\begin{figure*}[!htb]\centering
	\includegraphics[width=5.3in]{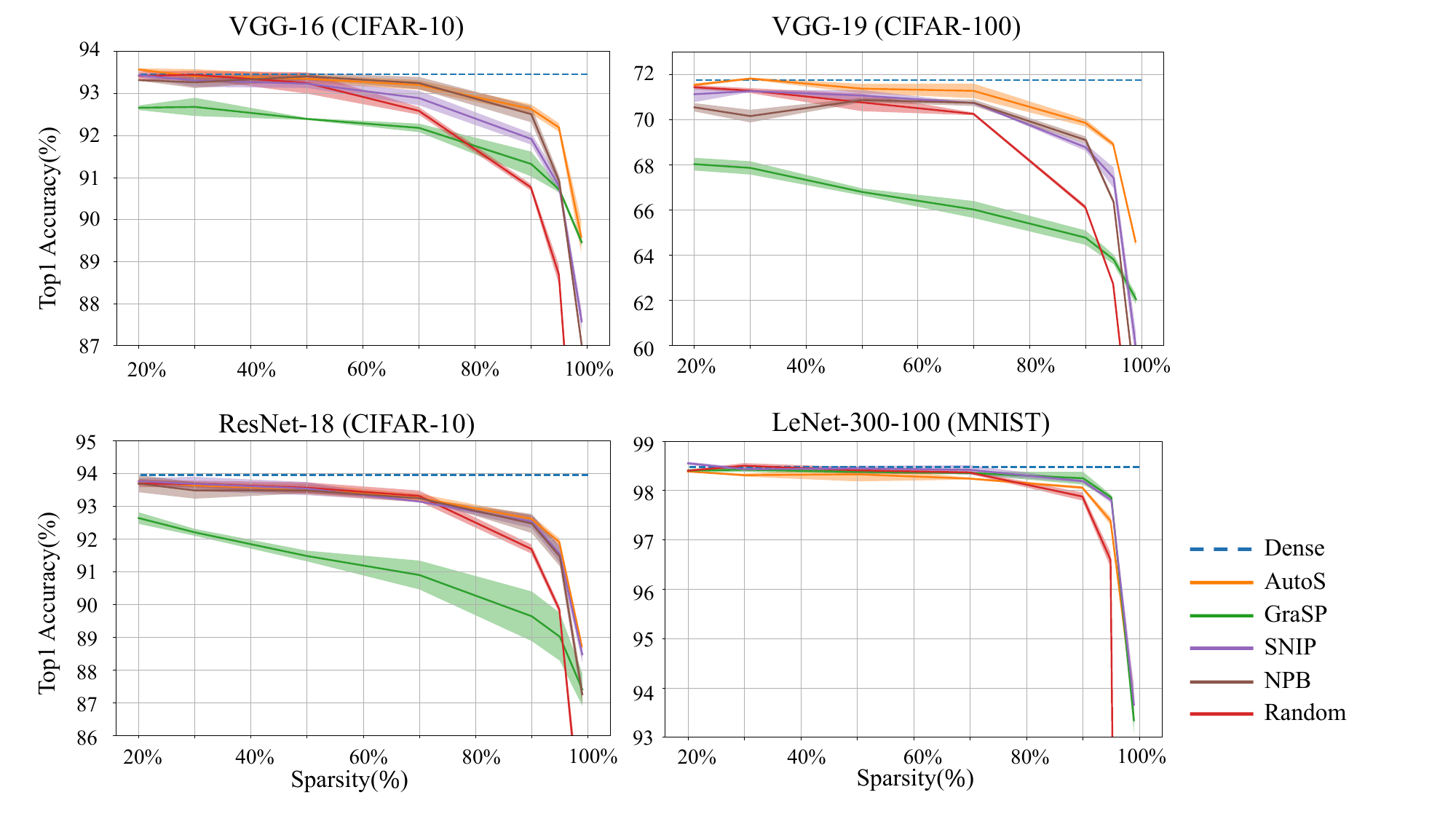}
	\caption{Accuracy of different PaI methods prune to various sparsities.}
	\label{fig_3}
\end{figure*}

In this section, we evaluate the baseline PaI methods and our AutoS on CIFAR-10, CIFAR-100, MNIST and TinyImageNet using various network architectures. Figure \ref{fig_3} presents the performance comparison of SNIP, GraSP, random, and AutoS. For context, the performance of the dense (unpruned) model is also included.

\paragraph{Experiment Setup:} We used CIFAR-10 to train ResNet-18 and employed SNIP to perform a one-time IRP with 20 iterations to build the dataset. This dataset was then used to train AutoS with ResNet-18, which was subsequently used to prune various networks. Full experimental details, including network architecture and hyperparameters, can be found in the Appendix \ref{data_gen} and \ref{pruning_details}.

\begin{figure*}[!htb]\centering
	\includegraphics[width=5 in]{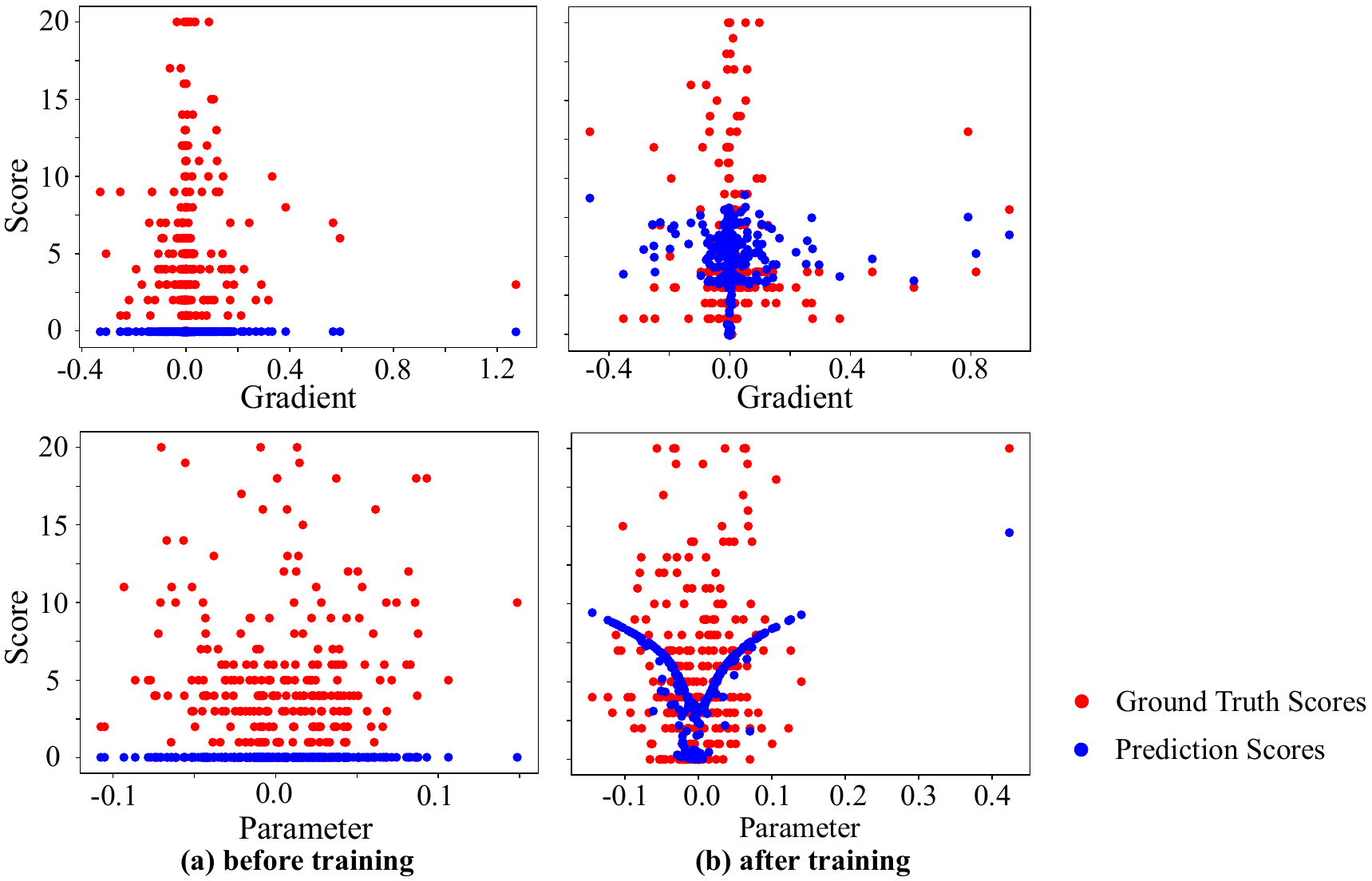}
	\caption{Visualization of score prediction. (a) is the prediction results before AutoS training, (b) is the prediction results after AutoS training.}
	\label{fig_4}
\end{figure*}

\paragraph{Analysis:} In Figure \ref{fig_3}, on VGG-16, AutoS outperforms SNIP by 2\% accuracy on CIFAR-10. Similarly, on VGG-19, our method achieves 2\% higher accuracy than SNIP. On ResNet-18, our method achieves 0.8\% higher accuracy than SNIP at 95\% sparsity (where only 5\% of parameters remain) and 0.5\% higher accuracy at 99\% sparsity. This proves that our method is effective for PaI. Notably, the above AutoS is trained with the dataset created by one-time IRP on ResNet-18 (CIFAR10). It demonstrates our model can be generalized well on unseen networks and only one-time IRP is needed. In our method, GraSP is strange at low-level sparsity, but it can achieve higher performance than SNIP and NPB at high-level sparsity, which is consistent with the evaluation in GraSP \citep{wang2020picking}. We also conducted AutoS on ResNet-18 with TinyImageNet in Table \ref{table_1} to test the robustness of the large-scale dataset. AutoS performs better than SNIP and GraSP at all four sparsities. Different from the above small-scale datasets, the NPB outperforms AutoS, especially at 70\% and 90\% sparsity. When the pruning ratio is 30\% or 50\%, AutoS is close to the NPB. This approves the effectiveness of AutoS on large-scale dataset.

\begin{table}
  \caption{Acccuracy of different PaI methods on ResNet-18 (TingImageNet).}
  \label{table_1}
  \centering
  \begin{tabular}{ccccc}
    \toprule
    	\multirow{2}{*}{Methods}  & \multicolumn{4}{c}{Sparsity}   \\
    	\cmidrule(r){2-5}
    
          &   30\% & 50\% & 70\% & 90\% \\
    	\midrule
			SNIP & 59.88$\pm$0.52  &  58.34$\pm$0.35 & 57.45$\pm$0.22  & 53.16$\pm$0.13  \\
			GraSP  & 57.83$\pm$0.65  &  56.54$\pm$0.42 & 55.43$\pm$0.17  &  54.43$\pm$0.09 \\
			NPB  &  60.35$\pm$0.35 & 59.35$\pm$0.34  &  58.70$\pm$0.30 & 56.4$\pm$0.22  \\
			AutoS & 60.02$\pm$0.23  & 59.23$\pm$0.27  & 58.05$\pm$0.20  & 55.32$\pm$0.15  \\
    \bottomrule
  \end{tabular}
\end{table}

\paragraph{Prediction Visualization:} To better understand what our AutoS predicts, we visualize the ground truth surviving scores and the predictions in Figure \ref{fig_4}. From Figure \ref{fig_4} (b), we observe that AutoS learns from the iterative pruning process that parameter values have a strong relationship with parameter scores: higher absolute parameter values correspond to relatively higher scores. Furthermore, gradients show a less significant relationship with parameter scores. This is evident as only a small number of points with high gradients have high scores, and many gradients close to 0 still have large scores. The ground truth dataset in Figure \ref{fig_4} appears more disordered when considering these two features. A clearer understanding can be gained from the ablation experiments discussed in Section \ref{abla}.

\section{Ablations}
\label{abla}

As our method uses a network to prune before training, we evaluate its performance across different datasets, backbone networks, and hyperparameters in this section. For better comparison, the default AutoS is presented in several tables, which is trained with \textbf{ResNet-18 (CIFAR-10)}, \textbf{SNIP-iter} and \textbf{20-iter}. We annotate this experiment in bold font. All the ablation implementation details are in Appendix \ref{abla_detail}.

%%%%%%%%%%% input
\subsection{Different Inputs}
\label{input}

To evaluate the effectiveness of different input features, we experiment with three types of inputs: 1) \textit{Only Param}: only use parameters as the input; 2) \textit{Only Grad}: only use gradient as the input; 3) \textit{Param + Grad}: combining both parameters and gradients as the input. We tested these approaches using ResNet-18 as our AutoS model and pruned ResNet-18 on CIFAR-10. The AutoS dataset is created using iterative SNIP (SNIP-iter). The results are shown in Table \ref{table_2}. The Only Para approach and the Only Grad approach drop a little at high level sparsity compared to the Param + Grad, with approximately $1\%$ accuracy lower. This suggests that initial parameter and gradient both contribute to PaI and inputting more useful initial features into AutoS leads to higher accuracy.

\begin{table}[H]
    \centering
    \small
    \begin{minipage}{0.47\textwidth}
        \centering
        \caption{Different inputs of AutoS to prune ResNet-18 (CIFAR-10)}
        \label{table_2}
        \begin{tabular}{ccccc}
            \toprule
            Input  & 70\%  & 90\% & 95\% & 99\% \\
            \midrule
            Only Param  & 93.07 & 92.40 & 91.78 & 87.6 \\
            Only Grad & 93.27 & 92.53 & 91.70 & 87.99 \\
            \textbf{Param + Grad}  & \textbf{93.09} & \textbf{92.63} & \textbf{92.00} & \textbf{88.78} \\
            \bottomrule
        \end{tabular}
    \end{minipage}%
    \hfill
    \begin{minipage}{0.47\textwidth}
        \centering
        \small
        \caption{LTH and SNIP-iter (dataset) to train AutoS to prune ResNet-18 (CIFAR-10)}
        \label{table_3}
        \begin{tabular}{ccccc}
            \toprule
            Method & 70\%  & 90\% & 95\% & 99\% \\
            \midrule
            LTH  & 93.41 & 92.09 & 89.96 & 79.15 \\
            \textbf{SNIP-iter}  & \textbf{93.09} & \textbf{92.63} & \textbf{92.00} & \textbf{88.78} \\
            \bottomrule
        \end{tabular}
    \end{minipage}
\end{table}

\begin{figure*}[!htb]\centering
	\includegraphics[width= 5 in]{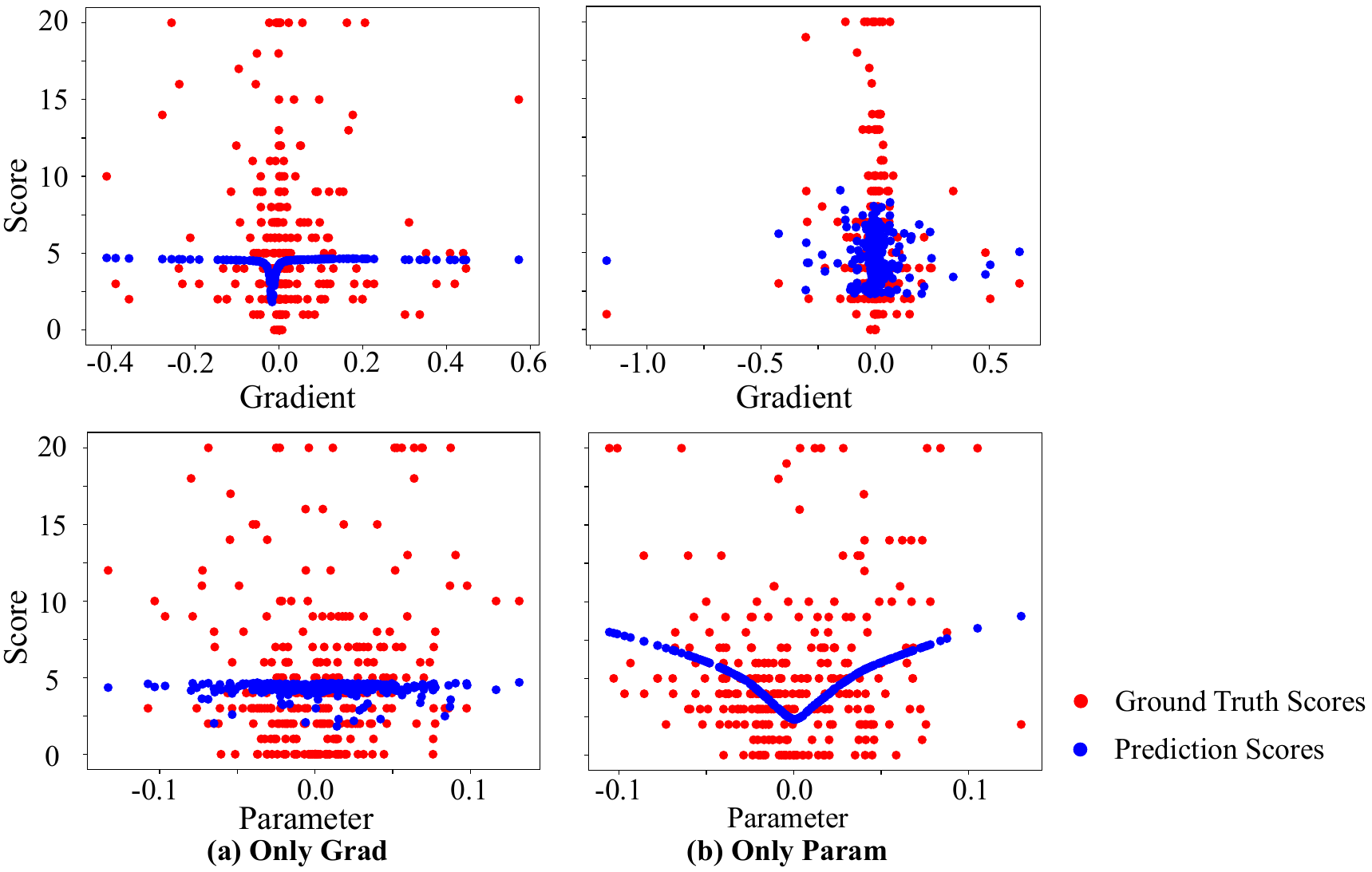}
	\caption{Visualization of ablation score prediction. (a) and (b) are the prediction results of Only Grad and Only Param.}
	\label{fig_5}
\end{figure*}

For further investigation of the relation between parameter and gradient inputs, we visualize the results of these two ablation networks in Figure \ref{fig_5}. It can be observed that when the feature is not input, there is no correspondence between it and the score. The results learned by the AutoS models with only Grad or Param as input are generally consistent with the results of Param + Grad shown in Figure \ref{fig_4}. However, the results obtained with individual inputs exhibit almost no noise and demonstrate a clear functional relationship with the corresponding parameters. With only Grad as input, the closer the absolute value of the gradient is to zero, the lower the score; as the absolute value of the gradient increases, the score rapidly reaches its peak and remains unchanged with further increases in gradient magnitude. This result indicates that as long as there is a gradient, the corresponding parameter is effective. This finding is consistent with other studies emphasizing the importance of parameter nodes \citep{pham2024towards} and parameter information transmission size \citep{vysogorets2023connectivity}. Similarly, for the model with only Param as input, the score increases as the absolute value of the parameter increases. This also reflects why the LTH \citep{frankle2018lottery}, which iteratively prunes based solely on parameter importance, can achieve an efficient subnetwork.

\begin{figure*}[!htb]\centering
	\includegraphics[width=5 in]{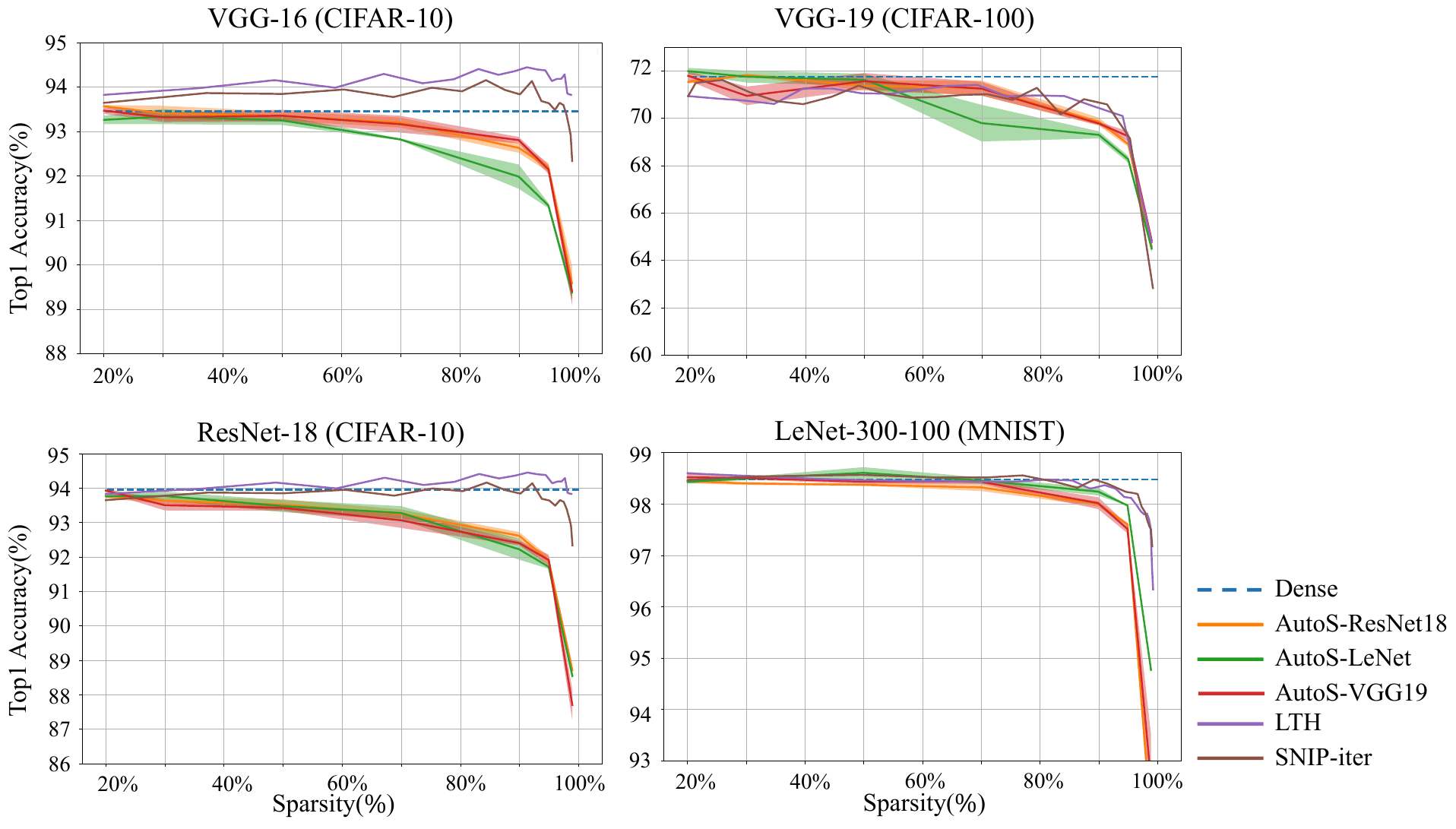}
	\caption{Accuracy of different ablation AutoS models and AutoS datasets $\mathcal{D}_{\mathcal{A}}$ to various sparsities.}
	\label{fig_6}
\end{figure*}

%%%%%%%%%%%%%%%%%% Dataset
\subsection{Different Datasets}
\label{dataset}

To evaluate the effectiveness of different IRP datasets in training AutoS, we have adopted four distinct aspects: 1) \textit{LTH or SNIP}: IRP like LTH or IRP using SNIP to generate dataset; 2) \textit{Model}: iterative pruning on different datasets (e.g., IRP on ResNet-18, VGG-19 et al.); 3) \textit{Iteration}: using 20, 40, or 100 iterations for generating the AutoS dataset; 4) \textit{Merge}: merging different datasets to train AutoS.

%%%%%%%%%%%%%%%% L or S
\paragraph{LTH or SNIP}
\label{LorS}

The different iterative pruning methods are to evaluate the best RIP method for the AutoS dataset. The results are shown in Table \ref{table_3}. The AutoS trained with SNIP-iter outperforms the LTH (magnitude) at sparsity levels beyond 90\%. We believe this is because LTH only involves a one-dimensional parameter feature, making it difficult for the network to effectively extract this feature. In contrast, SNIP-iter prunes parameters with the lowest gradient flow, capturing the relationship between parameters and gradients. This likely helps AutoS in effectively extracting the relation between the parameter and gradient from the dataset.

%%%%%%%%%%%%%%%%%%%%%  model

\begin{table}
  \caption{Different datasets to train AutoS and different models to implement AutoS to prune ResNet-18 (CIFAR-10)}
  \label{table_4}
  \centering
  \begin{tabular}{cccccc}
    \toprule
    \multicolumn{2}{c}{AutoS Dataset} & \multirow{2}{*}{AutoS Model} & \multicolumn{3}{c}{Sparsity}   \\
    \cmidrule(r){1-2} \cmidrule(r){4-6}
    Dataset     & Model  &   & 90\% & 95\% & 99\% \\
    \midrule
     \multirow{2}{*}{MNIST} & \multirow{2}{*}{LeNet-300-100}  &   ResNet-18    & 91.94 &  91.79 & 88.07 \\
     &   & VGG-19  & 91.88  & 91.60  &  88.00 \\
     \multirow{2}{*}{CIFAR-10}     & \multirow{2}{*}{ResNet-18} & \textbf{ResNet-18} & \textbf{92.63} & \textbf{92.00} & \textbf{88.78}   \\
     &  & VGG-19   & 92.46 & 91.83 & 88.24  \\
      \multirow{2}{*}{CIFAR-100} & \multirow{2}{*}{VGG-19} & ResNet-18  & 92.71 & 56.26 & 37.54         \\
      &  & VGG-19  & 92.41  & 91.63 & 19.40  \\
      \multirow{2}{*}{TinyImageNet}  & \multirow{2}{*}{ResNet-18}  & ResNet-18 & 92.55 & 91.55 & 64.04 \\
         &   & VGG-19  & 92.21  & 91.23  & 60.04 \\
    \bottomrule
  \end{tabular}
\end{table}

\paragraph{Model}
\label{model}
To evaluate which model is more effective for training AutoS, we use LeNet-300-100 (MNIST), ResNet-18 (CIFAR-10), and VGG-19 (CIFAR-100) to train AutoS for pruning ResNet-18 (CIFAR-10). The results are shown in Table \ref{table_4}. It can be seen that ResNet-18 (CIFAR-10) slightly outperforms LeNet-300-100 (MNIST) by approximately 0.2-0.7\%. However, the results trained with VGG-19 (CIFAR-100) drop significantly at 99\% sparsity. This suggests that the choice of model used to create the AutoS dataset can significantly impact the pruning performance of AutoS. LeNet-300-100 has about 60,000 parameters, ResNet-18 has 11.7 million parameters, and VGG-19 has 144 million parameters. Despite VGG-19 having the most parameters, it performs the worst. Given that the good performance of LeNet-300-100 on MNIST is similar to that of ResNet-18 on CIFAR-10, it suggests that model size may not significantly affect the AutoS dataset. When comparing the accuracy drop of AutoS using ResNet-18 on TinyImageNet, we conjecture that this might be due to the model's suboptimal performance on its original dataset. For instance, VGG-19 achieves only 72\% accuracy on CIFAR-100, and ResNet-18 achieves 61\% on TinyImageNet. This indicates that these models are not ideal, and their parameters do not follow an ideal pattern to effectively guide other models. Consequently, the IRP from these models cannot adequately teach an ideal AutoS.

%%%%%%%%%%%% 

We also implemente AutoS with VGG-19 and trained with the above four models to prune ResNet-18 (CIFAR-10) models. The results are shown in Table 4. Since the input and output data have no special structure, different models have little influence on the pruning accuracy. We selected two models from Table \ref{table_4} to conduct a full set of experiments for a comprehensive investigation, following the procedure of the main experiment. One model is AutoS-LeNet, from MNIST-\textbf{LeNet}-300-100 (Dataset) and ResNet-18 (AutoS Model), to test the pruning effect on a small AutoS dataset. The other is AutoS-VGG19, from CIFAR-10-ResNet-18 and \textbf{VGG-19} (AutoS Model), to investigate the different AutoS model's effectiveness. The results are presented in Figure \ref{fig_6}, which also includes pruning results from other methods (LTH and SNIP-iter).

The performances of all AutoS methods are generally similar, with the exception that AutoS from LeNet-300-100 shows slightly lower performance at high sparsity levels compared to AutoS from ResNet-18 and VGG-19 on CIFAR-10 with VGG-16 (CIFAR-10) and VGG-19 (CIFAR-100). Additionally, the results from LeNet-300-100 for pruning LeNet-300-100 outperform those of the other two methods, suggesting that the MNIST dataset may be too small, potentially leading to overfitting in AutoS. These results further demonstrate the generalization capability of AutoS, as all three datasets were created using a one-time IRP. While there is still a performance gap between AutoS and iterative pruning for most datasets, the performance on VGG-19 (CIFAR-100) is very close to that of iterative pruning in Figure \ref{fig_6}, significantly outperforming other state-of-the-art methods, as shown in Figure \ref{fig_3}.

%%%%%%%%%%%%%%%%%%%%%  iteration

\begin{table}
	\caption{Different iteration to train AutoS to prune ResNet-18 (CIFAR-10)}
	\label{table_5}
	\centering
    \begin{tabular}{ccccc}
    \toprule
    \multirow{2}{*}{Iterative Times}  & \multicolumn{4}{c}{Sparsity}   \\
    \cmidrule(r){2-5}
    
          &   70\%  & 90\% & 95\% & 99\% \\
    \midrule
          40-iter   & 93.21 & 92.57 &  91.88 &  88.66 \\
          100-iter  & 93.24  & 92.20  &  91.87 & 88.45  \\
	\bottomrule	
    \end{tabular}
\end{table}

\begin{table}
	\caption{Different iteration to train AutoS to prune ResNet-18 (CIFAR-10)}
	\label{table_6}
	\centering
    \begin{tabular}{ccccc}
    \toprule
    	\multirow{2}{*}{Merge}  & \multicolumn{4}{c}{Sparsity}   \\
    	\cmidrule(r){2-5}
    
          &   70\%  & 90\% & 95\% & 99\% \\
    	\midrule
          20+40+100      & 93.17 & 92.74 & \underline{47.56} & \underline{10.00} \\
          3*20   & 93.17 & 92.50 & 91.73 & 88.39 \\
          LTH+SNIP    & 92.90  & 92.59  &  91.70 & 88.45  \\
	\bottomrule	
    \end{tabular}
\end{table}

\paragraph{Iteration}
\label{iter}
We further investigate the effectiveness of fine-grained parameter scores by training AutoS with different pruning iterations. The results are shown in Table \ref{table_5}. AutoS trained with more fine-grained datasets exhibit a slight performance drop at sparsity levels from 90\% to 99\%. This decline may be due to the scores being too densely clustered, making it difficult for AutoS to effectively distinguish between them.

%%%%%%%%%%%%%%%%%%%%%  merge

\paragraph{Merge}
\label{merge}

We further merge different datasets to train AutoS to test whether a larger dataset can improve its performance. We try three approaches to merge datasets: 1) \textit{3*20}: merging three different 20-iteration datasets into one dataset (i.e., three-times IRP with different initial parameter); 2) \textit{20+40+100}: merging 20, 40, and 100 iterations into one dataset; 3) \textit{LTH + SNIP}: merging LTH and SNIP-iter datasets into one.

The results can be found in Table \ref{table_6}. The results of the 3*20 and Mag+SNIP merges do not show significant changes. Although the subnetworks from IRP are not the same \cite{frankle2018lottery}, this doesn't influence much for AutoS. The 20+40+100 merge experiences a significant drop at 95\% sparsity and fails to work at 99\% sparsity. We conjecture this might be because merging coarse-grained datasets with fine-grained datasets creates ambiguities. For example, in the 20-iteration dataset, 5\% of parameters might be scored as 1 (the score will be normalized to [0,1]), while in the 100-iteration dataset, only 1\% of parameters might be scored as 1, with the next 4\% scored as 0.99, 0.98, 0.97, and 0.96, respectively. This discrepancy makes the merged dataset "dirty", leading to AutoS's inability to learn effective predictions for fine-grained parameters, resulting in weak performance at high sparsity levels. Merging LTH and SNIP also has little influence on the pruning accuracy. Although these two methods use different criteria.

\section{Limitations}
\label{limit}

Our method demonstrates promising results in parameter importance (PaI), with its effectiveness confirmed through extensive experiments, including ablation studies. However, due to computational limitations, our experiments were primarily conducted on the MNIST, CIFAR-10, CIFAR-100 and TinyImageNet, utilizing models such as LeNet-300-100, ResNet-18, VGG-16 and VGG-19. As a result, the performance on these datasets may not fully represent the method's performance on other datasets. Furthermore, for some experimental results, we can only provide an experiential analysis based on our observations. The true underlying reasons require further experiments to demonstrate.

\section{Discussion}
\label{dissc}

Our novel PaI method leverages a neural network to learn effective criteria from iterative pruning. Although this method surpasses some state-of-the-art baselines, it still exhibits a performance gap relative to iterative pruning, particularly at high levels of sparsity in Figure \ref{fig_6}. From our comprehensive experiments, we infer three main aspects:

\subsection{Datadriven for PaI}
\paragraph{The state of the art} We propose using a neural network to learn the PaI criterion from iterative pruning, and our methods achieve state-of-the-art performance across most models. These results demonstrate the effectiveness of the surviving score from IRP and the generalization capabilities learned from this score. While previous methods have explored subnetworks obtained through LTH in various ways \citep{morcos2019one, chen2020lottery, frankle2019stabilizing}, our approach of utilizing these subnetworks to guide initialization pruning is novel and is being proposed for the first time in this work.

\paragraph{Network learning results} The results predicted by AutoS using only gradients or only parameters as input are astonishing, as they produce nearly perfect functional curves. This consistency with traditional manually designed pruning methods, viewed through the lens of neural networks' autonomous learning, further validates their effectiveness. However, given the complexity of the dataset, manually designing such curves is nearly impossible. Additionally, LTH intuitively assesses the importance of parameters based on their absolute values in a linear manner. While this approach aligns with network predictions, it offers limited insights into the mechanisms of network pruning. In contrast, our network predicts results that approximate a logarithmic curve. If these predictions can be theoretically validated, it would significantly advance our understanding and exploration of neural networks.

\subsection{Potential Improvements}

\paragraph{Datasets Quality} Our experiments demonstrate that a poor initial network fails to learn an ideal data mapping relationship, which in turn makes it difficult for the AutoS dataset generated through IRP to train an effective AutoS model. Consequently, there is still room for improvement in the performance of ResNet-18 on CIFAR-10 (with an accuracy of 92\%). If a more optimal dataset were used for IRP, it is likely that the resulting AutoS would achieve even better performance.

\paragraph{Superior Input Feature} Our experiments demonstrate that combining initial parameters with gradients yields better performance. Incorporating additional useful initial features as inputs could further enhance the accuracy of initialization pruning. For example, as noted in \citep{frankle2019stabilizing}, rewinding to the initial state often struggles to perform well in deeper networks, but rewinding the parameters at the 10th iteration of IRP has been shown to achieve further improvements. Additionally, the network's topology can also influence the model's accuracy. However, since these features are typically obtained through training, understanding how to relate them to the initial features and integrate them as inputs for AutoS could be an effective way to improve pruning accuracy.

\section{Conclusion}
\label{conc}

We introduce a novel PaI method that leverages neural networks to learn from the superior strategies of iterative pruning, thereby enhancing pruning effectiveness. By obtaining datasets once from iterative pruning processes and employing these in training our PaI network, AutoS, we have achieved more precise results than several mainstream PaI methods. Furthermore, AutoS can generalize well to unknown networks without additional training. Our experiments validate the viability of deriving PaI strategies from iterative pruning, demonstrating significant improvements in pruning accuracy. Additionally, our experiments suggest that future enhancements in dataset generation, feature input, and network architecture could further improve PaI accuracy with AutoS. We are optimistic that our work will inspire and inform subsequent research in this field.

\newpage

\bibliography{main}

@inproceedings{lee2018snip,
  title={Snip: Single-shot network pruning based on connection sensitivity},
  author={Lee, Namhoon and Ajanthan, Thalaiyasingam and Torr, Philip HS},
  booktitle={International Conference on Learning Representations},
  year={2018}
}

@article{lecun1989optimal,
  title={Optimal brain damage},
  author={LeCun, Yann and Denker, John and Solla, Sara},
  journal={Advances in neural information processing systems},
  volume={2},
  year={1989}
}

@article{reed1993pruning,
  title={Pruning algorithms-a survey},
  author={Reed, Russell},
  journal={IEEE transactions on Neural Networks},
  volume={4},
  number={5},
  pages={740--747},
  year={1993},
  publisher={IEEE}
}

@article{blalock2020state,
  title={What is the state of neural network pruning?},
  author={Blalock, Davis and Gonzalez Ortiz, Jose Javier and Frankle, Jonathan and Guttag, John},
  journal={Proceedings of machine learning and systems},
  volume={2},
  pages={129--146},
  year={2020}
}

@article{zhu2017prune,
  title={To prune, or not to prune: exploring the efficacy of pruning for model compression},
  author={Zhu, Michael and Gupta, Suyog},
  journal={arXiv preprint arXiv:1710.01878},
  year={2017}
}

@article{gale2019state,
  title={The state of sparsity in deep neural networks},
  author={Gale, Trevor and Elsen, Erich and Hooker, Sara},
  journal={arXiv preprint arXiv:1902.09574},
  year={2019}
}

@article{han2015learning,
  title={Learning both weights and connections for efficient neural network},
  author={Han, Song and Pool, Jeff and Tran, John and Dally, William},
  journal={Advances in neural information processing systems},
  volume={28},
  year={2015}
}

@article{brown2020language,
  title={Language models are few-shot learners},
  author={Brown, Tom and Mann, Benjamin and Ryder, Nick and Subbiah, Melanie and Kaplan, Jared D and Dhariwal, Prafulla and Neelakantan, Arvind and Shyam, Pranav and Sastry, Girish and Askell, Amanda and others},
  journal={Advances in neural information processing systems},
  volume={33},
  pages={1877--1901},
  year={2020}
}

@article{frankle2018lottery,
  title={The lottery ticket hypothesis: Finding sparse, trainable neural networks},
  author={Frankle, Jonathan and Carbin, Michael},
  journal={International Conference on Learning Representations},
  year={2019}
}

@article{wang2020picking,
  title={Picking winning tickets before training by preserving gradient flow},
  author={Wang, Chaoqi and Zhang, Guodong and Grosse, Roger},
  journal={International Conference on Learning Representations},
  year={2020}
}

@article{tanaka2020pruning,
  title={Pruning neural networks without any data by iteratively conserving synaptic flow},
  author={Tanaka, Hidenori and Kunin, Daniel and Yamins, Daniel L and Ganguli, Surya},
  journal={Advances in neural information processing systems},
  volume={33},
  pages={6377--6389},
  year={2020}
}

@article{hoang2023revisiting,
  title={Revisiting pruning at initialization through the lens of ramanujan graph},
  author={Hoang, Duc NM and Liu, Shiwei},
  journal={International Conference on Learning Representations},
  year={2023}
}

@article{frankle2020pruning,
  title={Pruning neural networks at initialization: Why are we missing the mark?},
  author={Frankle, Jonathan and Dziugaite, Gintare Karolina and Roy, Daniel M and Carbin, Michael},
  journal={International Conference on Learning Representations},
  year={2020}
}

@article{wimmer2023dimensionality,
  title={Dimensionality reduced training by pruning and freezing parts of a deep neural network: a survey},
  author={Wimmer, Paul and Mehnert, Jens and Condurache, Alexandru Paul},
  journal={Artificial Intelligence Review},
  volume={56},
  number={12},
  pages={14257--14295},
  year={2023},
  publisher={Springer}
}

@article{liu2018rethinking,
  title={Rethinking the value of network pruning},
  author={Liu, Zhuang and Sun, Mingjie and Zhou, Tinghui and Huang, Gao and Darrell, Trevor},
  journal={International Conference on Learning Representations},
  year={2019}
}

@article{li2017pruning,
  title={Pruning filters for efficient convnets,},
  author={Hao, Li and Asim Kadav and Igor Durdanovic and Hanan and Samet and Hans Peter Graf},
  journal={International Conference on Learning Representations},
  year={2017}
}

@inproceedings{mao2017exploring,
  title={Exploring the granularity of sparsity in convolutional neural networks},
  author={Mao, Huizi and Han, Song and Pool, Jeff and Li, Wenshuo and Liu, Xingyu and Wang, Yu and Dally, William J},
  booktitle={Proceedings of the IEEE Conference on Computer Vision and Pattern Recognition Workshops},
  pages={13--20},
  year={2017}
}

@article{mocanu2018scalable,
  title={Scalable training of artificial neural networks with adaptive sparse connectivity inspired by network science},
  author={Mocanu, Decebal Constantin and Mocanu, Elena and Stone, Peter and Nguyen, Phuong H and Gibescu, Madeleine and Liotta, Antonio},
  journal={Nature communications},
  volume={9},
  number={1},
  pages={2383},
  year={2018},
  publisher={Nature Publishing Group UK London}
}

@inproceedings{evci2020rigging,
  title={Rigging the lottery: Making all tickets winners},
  author={Evci, Utku and Gale, Trevor and Menick, Jacob and Castro, Pablo Samuel and Elsen, Erich},
  booktitle={International conference on machine learning},
  pages={2943--2952},
  year={2020},
}

@article{dong2017learning,
  title={Learning to prune deep neural networks via layer-wise optimal brain surgeon},
  author={Dong, Xin and Chen, Shangyu and Pan, Sinno},
  journal={Advances in neural information processing systems},
  volume={30},
  year={2017}
}

@article{molchanov2016pruning,
  title={Pruning convolutional neural networks for resource efficient inference},
  author={Molchanov, Pavlo and Tyree, Stephen and Karras, Tero and Aila, Timo and Kautz, Jan},
  journal={International Conference on Learning Representations},
  year={2016}
}

@inproceedings{wang2019eigendamage,
  title={Eigendamage: Structured pruning in the kronecker-factored eigenbasis},
  author={Wang, Chaoqi and Grosse, Roger and Fidler, Sanja and Zhang, Guodong},
  booktitle={International conference on machine learning},
  pages={6566--6575},
  year={2019},
  organization={PMLR}
}

@inproceedings{yu2018nisp,
  title={Nisp: Pruning networks using neuron importance score propagation},
  author={Yu, Ruichi and Li, Ang and Chen, Chun-Fu and Lai, Jui-Hsin and Morariu, Vlad I and Han, Xintong and Gao, Mingfei and Lin, Ching-Yung and Davis, Larry S},
  booktitle={Proceedings of the IEEE conference on computer vision and pattern recognition},
  pages={9194--9203},
  year={2018}
}

@article{guo2016dynamic,
  title={Dynamic network surgery for efficient dnns},
  author={Guo, Yiwen and Yao, Anbang and Chen, Yurong},
  journal={Advances in neural information processing systems},
  volume={29},
  year={2016}
}

@article{liu2022unreasonable,
  title={The unreasonable effectiveness of random pruning: Return of the most naive baseline for sparse training},
  author={Liu, Shiwei and Chen, Tianlong and Chen, Xiaohan and Shen, Li and Mocanu, Decebal Constantin and Wang, Zhangyang and Pechenizkiy, Mykola},
  journal={arXiv preprint arXiv:2202.02643},
  year={2022}
}

@article{verdenius2020pruning,
  title={Pruning via iterative ranking of sensitivity statistics},
  author={Verdenius, Stijn and Stol, Maarten and Forr{\'e}, Patrick},
  journal={arXiv preprint arXiv:2006.00896},
  year={2020}
}

@inproceedings{he2016deep,
  title={Deep residual learning for image recognition},
  author={He, Kaiming and Zhang, Xiangyu and Ren, Shaoqing and Sun, Jian},
  booktitle={Proceedings of the IEEE conference on computer vision and pattern recognition},
  pages={770--778},
  year={2016}
}

@article{pham2024towards,
  title={Towards data-agnostic pruning at initialization: what makes a good sparse mask?},
  author={Pham, Hoang and Liu, Shiwei and Xiang, Lichuan and Le, Dung and Wen, Hongkai and Tran-Thanh, Long and others},
  journal={Advances in Neural Information Processing Systems},
  volume={36},
  year={2024}
}

@article{vysogorets2023connectivity,
  title={Connectivity matters: Neural network pruning through the lens of effective sparsity},
  author={Vysogorets, Artem and Kempe, Julia},
  journal={Journal of Machine Learning Research},
  volume={24},
  number={99},
  pages={1--23},
  year={2023}
}

@article{morcos2019one,
  title={One ticket to win them all: generalizing lottery ticket initializations across datasets and optimizers},
  author={Morcos, Ari and Yu, Haonan and Paganini, Michela and Tian, Yuandong},
  journal={Advances in neural information processing systems},
  volume={32},
  year={2019}
}

@article{chen2020lottery,
  title={The lottery ticket hypothesis for pre-trained bert networks},
  author={Chen, Tianlong and Frankle, Jonathan and Chang, Shiyu and Liu, Sijia and Zhang, Yang and Wang, Zhangyang and Carbin, Michael},
  journal={Advances in neural information processing systems},
  volume={33},
  pages={15834--15846},
  year={2020}
}

@article{frankle2019stabilizing,
  title={Stabilizing the lottery ticket hypothesis},
  author={Frankle, Jonathan and Dziugaite, Gintare Karolina and Roy, Daniel M and Carbin, Michael},
  journal={arXiv preprint arXiv:1903.01611},
  year={2019}
}

\begin{appendix}
\appendix

\section{Dataset Gneration Details}
\label{data_gen}

We use the following combinations of networks and datasets for pruning.

\begin{table}[h]
	\label{table_networks}
	\centering
    \begin{tabular}{ccc}
    \toprule
    Network & Dataset & Appears  \\
    \midrule
    	  LeNet-300-100 & MNIST & Main Body\\
    	  ResNet-18 \& VGG-16 & CIFAR-10 & Main Body \& Appendix \\
    	  ResNet-18 \& VGG-19 & CIFAR-100 & Main Body \\
    	  ResNet-18 & TinyImageNet & Main Body\\
          
	\bottomrule	
    \end{tabular}
    
\end{table}

\subsection{Networks}

The network architectures are as follows:

\textbf{LeNet-300-100}: this is a fully-connected network designed for the MNIST dataset. The network consists of two hidden layers: the first hidden layer has 300 units, and the second hidden layer has 100 units. Both layers use ReLU activations.

\begin{table}[h]
	\label{lenet}
	\centering

    \begin{tabular}{cccccc}
    \toprule
    Module & Weight & Bias & BatchNorm & Activation \\
    \midrule
         
    	 Linear & [784, 300] & 300 & \ding{55} & RELU \\
    	 Linear & [300, 100] & 100 & \ding{55} & RELU \\
    	 Linear & [100, 10] & 10 & \ding{55} & - \\
    	
	\bottomrule	
    \end{tabular}
\end{table}
	
\textbf{ResNet-18}: this is a modified version where the first convolution layer has a filter size of 3x3, and the subsequent max-pooling layer has been eliminated. 

\begin{table}[H]
	\label{ResNet-18}
	\centering

    \begin{tabular}{ccccc}
    \toprule
    Module & Weight & Stride & BatchNorm & ReLU \\
    \midrule
         Conv & [3, 3, 3, 64] & 2 & \ding{51} & \ding{51}  \\
%    	 MaxPool & 3x3 & [64, 64] & 2 & \ding{55} & \ding{55s} \\
    	 Conv & [3, 3, 64, 64] & 1 & \ding{51} & \ding{51}  \\
    	 Conv & [3, 3, 64, 64] & 1 & \ding{51} & \ding{51}  \\
    	 \hline
    	 Conv & [3, 3, 64, 128] & 2 & \ding{51} & \ding{51}  \\
    	 Conv & [3, 3, 128, 128] & 1 & \ding{51} & \ding{51}  \\
    	 \hline
    	 Conv & [3, 3, 128, 256] & 2 & \ding{51} & \ding{51}  \\
    	 Conv & [3, 3, 256, 256] & 2 & \ding{51} & \ding{51}  \\
    	 \hline
    	 Conv & [3, 3, 256, 512] & 2 & \ding{51} & \ding{51}  \\
    	 Conv & [3, 3, 521, 521] & 2 & \ding{51} & \ding{51}  \\
    	 \hline
    	 Average Pooling & [1, 1, 512, 521] & 1 & \ding{55} & \ding{55}  \\
    	 Linear & [521, 1000] & - & \ding{55} & \ding{55}  \\

	\bottomrule	
    \end{tabular}
\end{table}

\textbf{VGG-16}: this is described by \cite{frankle2018lottery}, the architecture is as follows.

\begin{table}[H]
	
	\centering

    \begin{tabular}{cccccc}
    \toprule
    Module & Weight & Stride & Bias & BatchNorm & ReLU \\
    \midrule
         Conv & [3, 3, 3, 64] & [1, 1] & 64 & \ding{51} & \ding{51}  \\
%    	 MaxPool & 3x3 & [64, 64] & 2 & \ding{55} & \ding{55s} \\
    	 Conv & [3, 3, 64, 64] & [1, 1] & 64 & \ding{51} & \ding{51}  \\
    	 MaxPool & - & [2, 2] & - & \ding{55} & \ding{55} \\
    	 \hline
    	 Conv & [3, 3, 64, 128] & [1, 1] & 128 & \ding{51} & \ding{51} \\
        Conv & [3, 3, 128, 128] & [1, 1] & 128 & \ding{51} & \ding{51} \\
        MaxPool & - & [2, 2] & - & \ding{55} & \ding{55} \\
        \hline
        Conv & [3, 3, 128, 256] & [1, 1] & 256 & \ding{51} & \ding{51} \\
        Conv & [3, 3, 256, 256] & [1, 1] & 256 & \ding{51} & \ding{51} \\
        Conv & [3, 3, 256, 256] & [1, 1] & 256 & \ding{51} & \ding{51} \\
        MaxPool & - & [2, 2] & - & \ding{55} & \ding{55} \\
        \hline
        Conv & [3, 3, 256, 512] & [1, 1] & 512 & \ding{51} & \ding{51} \\
        Conv & [3, 3, 512, 512] & [1, 1] & 512 & \ding{51} & \ding{51} \\
        Conv & [3, 3, 512, 512] & [1, 1] & 512 & \ding{51} & \ding{51} \\
        MaxPool & - & [2, 2] & - & \ding{55} & \ding{55} \\
        \hline
        Conv & [3, 3, 512, 512] & [1, 1] & 512 & \ding{51} & \ding{51} \\
        Conv & [3, 3, 512, 512] & [1, 1] & 512 & \ding{51} & \ding{51} \\
        Conv & [3, 3, 512, 512] & [1, 1] & 512 & \ding{51} & \ding{51} \\
        MaxPool & - & [2, 2] & - & \ding{55} & \ding{55} \\
        \hline
        Linear & [512 * 7 * 7, 4096] & - & 4096 & \ding{55} & \ding{51} \\
        Linear & [4096, 4096] & - & 4096 & \ding{55} & \ding{51} \\
        Linear & [4096, 1000] & - & 1000 & \ding{55} & \ding{55} \\

	\bottomrule	
    \end{tabular}
\end{table}

\textbf{VGG-19}: this is the plus version of VGG-16. The first two layers have 64 channels followed by 2x2 max pooling, the next two layers have 128 channels followed by 2x2 max pooling; the next four layers have 256 channels followed by 2x2 max pooling; the next four layers have 512 channels followed by 2x2 max pooling; the next four layers have 512 channels. Each channel uses 3x3 convolutional filters. VGG-19 has batch normalization before each RELU activation.

\begin{table}[H]
	
	\centering

    \begin{tabular}{cccccc}
    \toprule
    Module & Weight & Stride & Bias & BatchNorm & ReLU \\
    \midrule
		Conv & [3, 3, 3, 64] & [1, 1] & [64] & \ding{51} & \ding{51} \\
        Conv & [3, 3, 64, 64] & [1, 1] & [64] & \ding{51} & \ding{51} \\
        MaxPool & - & [2, 2] & - & \ding{55} & \ding{55} \\
        \hline
        Conv & [3, 3, 64, 128] & [1, 1] & [128] & \ding{51} & \ding{51} \\
        Conv & [3, 3, 128, 128] & [1, 1] & [128] & \ding{51} & \ding{51} \\
        MaxPool & - & [2, 2] & - & \ding{55} & \ding{55} \\
        \hline
        Conv & [3, 3, 128, 256] & [1, 1] & [256] & \ding{51} & \ding{51} \\
        Conv & [3, 3, 256, 256] & [1, 1] & [256] & \ding{51} & \ding{51} \\
        Conv & [3, 3, 256, 256] & [1, 1] & [256] & \ding{51} & \ding{51} \\
        Conv & [3, 3, 256, 256] & [1, 1] & [256] & \ding{51} & \ding{51} \\
        MaxPool & - & [2, 2] & - & \ding{55} & \ding{55} \\
        \hline
        Conv & [3, 3, 256, 512] & [1, 1] & [512] & \ding{51} & \ding{51} \\
        Conv & [3, 3, 512, 512] & [1, 1] & [512] & \ding{51} & \ding{51} \\
        Conv & [3, 3, 512, 512] & [1, 1] & [512] & \ding{51} & \ding{51} \\
        Conv & [3, 3, 512, 512] & [1, 1] & [512] & \ding{51} & \ding{51} \\
        MaxPool & - & [2, 2] & - & \ding{55} & \ding{55} \\
        \hline
        Conv & [3, 3, 512, 512] & [1, 1] & [512] & \ding{51} & \ding{51} \\
        Conv & [3, 3, 512, 512] & [1, 1] & [512] & \ding{51} & \ding{51} \\
        Conv & [3, 3, 512, 512] & [1, 1] & [512] & \ding{51} & \ding{51} \\
        Conv & [3, 3, 512, 512] & [1, 1] & [512] & \ding{51} & \ding{51} \\
        MaxPool & - & [2, 2] & - & \ding{55} & \ding{55} \\
        \hline
        Linear & [512 * 7 * 7, 4096] & - & [4096] & \ding{55} & \ding{51} \\
        Linear & [4096, 4096] & - & [4096] & \ding{55} & \ding{51} \\
        Linear & [4096, 1000] & - & [1000] & \ding{55} & \ding{55} \\
 		\bottomrule	
    \end{tabular}
\end{table}

\subsection{Datasets}
\begin{itemize}
	\item CIFAR-10 consists of 60,000 color images, each with a resolution of 32x32 pixels. These images are categorized into 10 different classes, with 6,000 images per class.

	\item CIFAR-100 consists of 60,000 color images, each with a resolution of 32x32 pixels, divided into 100 different classes. Each class contains 600 images, providing a more fine-grained classification challenge compared to its counterpart, CIFAR-10.
	\item MNIST consists of 70,000 grayscale images of handwritten digits, each with a resolution of 28x28 pixels. It includes 60,000 training images and 10,000 test images, with labels from 0 to 9, making it ideal for evaluating classification algorithms.
	\item The TinyImageNet dataset is a subset of the ImageNet database, consisting of 200 classes with 500 training images and 50 validation images per class, each resized to 64x64 pixels. 
\end{itemize}

%%%%%%%%%%%%%%%%%%%%%%%%%%%%%
\subsection{Training and Pruning Details}

Following is the AutoS dataset generation. All the final sparsity is 0.01 (1\% parameter is remained). LTH \citep{frankle2018lottery} and SNIP \citep{lee2018snip} are based on the official code of Synflow\citep{tanaka2020pruning}\footnote{\url{https://github.com/ganguli-lab/Synaptic-Flow}}.

\begin{table}[H]
	
	\centering
	\tiny
    \begin{tabular}{cccccccccc}
    \toprule
    Network & Dataset & Post-Epochs & Opt. & Batch & LR & Drop Rate \& Epochs & Weight Decay  & Prune Iter. \\
    \midrule
    	  LeNet-300-100 & MNIST & 100 & Adam & 128 & 1.2e-3 & 0.2; 60, 120 & 5e-4  & 20 \\
    	  ResNet-18  & CIFAR-10 & 160 & momentum & 128 & 0.01 & 0.2; 60, 120 & 5e-4 & 20, 40, 100 \\
    	  VGG-19 & CIFAR-100 & 160 & momentum & 128 & 0.1 & 0.1; 60, 120 & 1e-4 & 20 \\
    	  ResNet-18 & TinyImageNet & 300 & SGD & 128 & 0.1 & 0.1; 150, 225 & 1e-4 & 20 \\
    	  
	\bottomrule	
    \end{tabular}
\end{table}

%%%%%%%%%%%%%%%%%%%%%%%%%%%%%

\section{AutoS Training Details}
\label{train_autos}

\subsection{Pseudocode}
\label{pseudocode}

The dataset generation process is the Algorithm \ref{code_datagen}.
\begin{algorithm}
\caption{AutoS Dataset ($\mathcal{D}_\mathcal{A}$) generation}
\label{code_datagen}
\begin{algorithmic}[1]
\Require  \\
Dataset $\mathcal{D}$: Pruning Dataset; \\
Network: a network parameterized by $\theta \in \mathbb{R}^k$; \\
Mask: corresponding pruning masks $m \in {\{ 0, 1 \}}^k$; \\
Target Sparsity $\epsilon$: desired sparsity level $\epsilon \in [0,1]$;\\
Pruning Criteria: a pruning criteria $C$ for iterative pruning; \\
iteration: pruning iterative times $N$.

\renewcommand{\algorithmicrequire}{\textbf{Input:}}
\Require $\theta_0$: network initialize; $m_0$: mask initialize
	\State \textit{model\_eval}: $g_0(\mathcal{D}) \leftarrow (\theta_0, \mathcal{D})$; \hfill model eval for initial gradient
	\For{$i < N$}: \hfill pruning iteration
	\State \quad $\theta_{i+1} \leftarrow$ \textit{model\_train($\theta_{i}(\theta_0, m_i), \mathcal{D}$)}; \hfill training
	\State \quad $\epsilon_{i+1} = (\epsilon, i)$ \hfill current sparsity
	\State \quad $m_{i+1}$ $\leftarrow$ $C(\theta_{i+1}, \epsilon_{i+1})$ \hfill pruning
	\State \quad $S_{surv}^{gt}$ \ $+= m_{i+1}$ \hfill score
	\State \quad $rewind$ \hfill rewind the parameter to $\theta_0$
	\EndFor

\renewcommand{\algorithmicrequire}{\textbf{Output:}}
\Require {$\mathcal{D}_\mathcal{A}$: \textbf{data}: initial gradient $g_0(\mathcal{D})$ and initial parameter $\theta_0$; \textbf{label}: importance score $S_{surv}^{gt}$}
\end{algorithmic}
\end{algorithm}

\subsection{AutoS Networks and Training}
We use ResNet-18 and VGG-19 as the backbone of AutoS. We add MLP layers in front of the major backbone to encode input features as the following Table \ref{autos_network}.  AutoS in Table \ref{table_1}, \ref{table_2},  \ref{table_4}, \ref{table_5}, and \ref{table_6} are implemented with ResNet-18.

%%%%%%%%%%%%%%% resnet
\begin{table}[H]
	
	\centering
    \caption{Network backbone of AutoS}
	\label{autos_network}
    \begin{tabular}{cccccc}
    \toprule
   AutoS Model & Module & Weight & Bias & BatchNorm & ReLU \\
    \midrule
    	\multirow{3}{*}{ResNet-18} & Linear & [2, 64] & 64 & \ding{51} & \ding{51} \\ 
    	& Linear & [64, 32 $\times$ 32 $\times$ 3] & 32 $\times$ 32 $\times$ 3 & \ding{51} & \ding{51} \\
    	\cmidrule(r){3-6}
    	& Backbone &\multicolumn{4}{c}{ResNet-18} \\
    	\cmidrule(r){3-6}
    	& Linear & [32, 1] & [32, 1] &  &  \ding{51} \\
    	\midrule
    	
    	\multirow{3}{*}{VGG-19} & Linear & [2, 64] & 64 & \ding{51} & \ding{51} \\ 
    	& Linear & [64, 64 $\times$ 64] & 32 $\times$ 32 $\times$ 3 & \ding{51} & \ding{51} \\
    	& Linear & [64 $\times$ 64, 64 $\times$ 64 $\times$ 64] & 64 $\times$ 64 $\times$ 64 & \ding{51} & \ding{51} \\
    	\cmidrule(r){3-6}
    	& Backbone & \multicolumn{4}{c}{VGG-19} \\ 
    	\cmidrule(r){3-6}
    	& Linear & [64, 1] & [64, 1] & \ding{51} &  \ding{51} \\
	\bottomrule	
    \end{tabular}
\end{table}

\subsection{Learning Rate and Batch Size Ablation}
 The learning rate is 0.01, batch size is 1024, epochs is 10, optimizer is Adam, loss with MSE. The ablation of batch size and learning rate are tested as following Table \ref{batch_lr_ablation}.

\begin{table}[H]
    \centering
    \caption{Learning rate ablation experiments}
	\label{batch_lr_ablation}
    \begin{tabular}{cccc}
    \toprule
    Batch & LR & 95\% & 99\% \\
    \midrule
    128 & 0.01 & 91.65 & 54.21 \\
    256 & 0.01 & 91.72 & 79.59 \\
    512 & 0.01 & 91.70 & 55.70 \\
    1024 & 0.01 & 91.90 & 79.20 \\
    2048 & 0.01 & 91.75 & 71.28 \\
    4096 & 0.01 & 82.26 & 53.49 \\ 
    4096 & 0.001 & 91.49 & 71.07 \\
    4096 & 0.0005 & 91.60 & 53.54 \\
    4096 & 0.0001 & 83.07 & 62.92 \\
     	  
	\bottomrule	
    \end{tabular}
\end{table}

\section{Pruning Details}
\label{pruning_details}

Following are all the pruning hyperparameters, all the pruning methods only pruning the neural network once after Kaiming Normal Initialization \citep{he2016deep}. AutoS in Table \ref{table_1},  \ref{table_2}, \ref{table_3}, \ref{table_4}, \ref{table_5}, and \ref{table_6}.

\begin{table}[H]
	\centering
	\tiny
    \begin{tabular}{cccccccccc}
    \toprule
    Network & Dataset & Post-Epochs & Opt. & Batch & LR & Drop Rate \& Epochs & Weight Decay \\
    \midrule
    	  LeNet-300-100 & MNIST & 100 & Adam & 128 & 1.2e-3 & 0.2; 60, 120 & 5e-4   \\
    	  ResNet-18  & CIFAR-10 & 160 & momentum & 128 & 0.01 & 0.2; 60, 120 & 5e-4  \\
    	  VGG-19 & CIFAR-100 & 160 & momentum & 128 & 0.1 & 0.1; 60, 120 & 1e-4 \\
    	  ResNet-18 & TinyImageNet & 300 & SGD & 128 & 0.1 & 0.1; 150, 225 & 1e-4  \\
    	  VGG-16 & CIFAR-10 & 160 & momentum & 128 & 0.1 & 0.1; 60, 120; & 1e-4 \\
    	  
	\bottomrule	
    \end{tabular}
\end{table}

\section{Ablation Details}
\label{abla_detail}

\subsection{Different Inputs}

The AutoS inputs experiments \textit{Only Param} and \textit{Only Grad} use the following network. The input channel is 1.

\begin{table}[H]
	
	\centering
    \begin{tabular}{cccccc}
    \toprule
   AutoS Model & Module & Weight & Bias & BatchNorm & ReLU \\
    \midrule
    	\multirow{3}{*}{ResNet-18} & Linear & [1, 64] & 64 & \ding{51} & \ding{51} \\ 
    	& Linear & [64, 32 $\times$ 32 $\times$ 3] & 32 $\times$ 32 $\times$ 3 & \ding{51} & \ding{51} \\
    	\cmidrule(r){3-6}
    	& Backbone &\multicolumn{4}{c}{Backbone ResNet-18} \\
    	\cmidrule(r){3-6}
    	& Linear & [32, 1] & [32, 1] &  &  \ding{51} \\
	\bottomrule	
    \end{tabular}
\end{table}

\subsection{Different Dataset}

The training experiments of ``LTH or SNIP", ``model", ``iteration" and ``merge" are the same as dataset generation and pruning in Appendix \ref{data_gen}. Here, the refine learning rate is 0.01.

\section{Experimental Resources}
\label{computer_resource}

All the experiments are run with two servers, the information is as follows:

\begin{table}[H]
	
	\centering
    \begin{tabular}{cccc}
    \toprule
   Sever & CPU & GPU & RAM \\
    \midrule
    	1 & Intel i9-14900K & 2$\times$3090, 24G & 128G  \\ 
    	2 & 2$\times$Intel Xeon 6226R & 8$\times$3090, 24G & 256G \\
    \bottomrule	
    \end{tabular}
\end{table}

The error bars in Figure \ref{fig_3} and \ref{fig_6} are running with seed 0,1,2.

\section{Consistency of surviving score}

We conducted supplementary experiments of the AutoS dataset obtained on the MNIST dataset to test the consistency of important scores (given that LTH is valid across different methods, the results on MNIST can represent results on other datasets).

Method: Use the same initial parameter of LeNet-300-100 on MNIST to create 2 AutoS datasets A and B. We use the IRP methods of Magnitude (Mag), Random (Rand), SNIP, and GraSP to generate A and B datasets. Assuming that we perform independent pruning on datasets A and B, obtaining importance scores X and Y respectively, we calculate the concordance correlation coefficient (CCC) to measure the consistency of the importance scores. The CCC formulation can be written as:

\begin{equation}
\label{equa_5}
\rho_c=\frac{2 \cdot \operatorname{Cov}(X, Y)}{\sigma_{X 2}+\sigma_{Y 2}+(\mu X-\mu Y)^2}
\end{equation}

The results are shown in Table \ref{table_ccc}, with values closer to 1 indicating greater consistency. From Table \ref{table_ccc}, we can see that the same initial parameters with higher CCC values, particularly in the SNIP \& SNIP and Mag \& Mag comparisons, exhibit greater consistency. This indicates that our importance scoring metric is robust to some extent. The results from LTH indicate that subnetworks are not unique, and different initial models can lead to different subnetworks, resulting in varying importance scores. Therefore, even though the CCC may not be very high, the criterion learned from one effective subnetwork can still be valuable for pruning.

\begin{table}[H]
	
	\centering
	\caption{Concordance Correlation Coefficient (CCC) of the Important Scores AutoS Dataset of Different Methods.}
	\label{table_ccc}
    \begin{tabular}{ccc}
    \toprule
          Methods & Same Init. & Different Init. \\
    	\midrule
    	  Random \& Random   & -0.00106  &  -0.00106   \\
    	  \textbf{SNIP \& SNIP}   &  \textbf{0.23642}  &   -0.04542  \\
    	  \textbf{Mag \& Mag}   &  \textbf{0.15582}  &   0.10285   \\
    	  GraSP \& GraSP  &  0.08059  &    0.04373  \\
    	  Random \& Mag &  0.00083  &   -0.00413  \\
    	  Random \& GraSP    &  0.00109  &  0.00259  \\
    	  Random \& SNIP  &  -0.00108    &  -0.00245  \\
    	  SNIP \& Mag    &  0.12314    &  0.03424  \\
    	  SNIP \& GraSP &  -0.07117   &  -0.04450  \\
    	  Mag \& GraSP    &  -0.05968   &  -0.01872  \\
	\bottomrule	
    \end{tabular}
\end{table}
\end{appendix}

\newpage

\end{document}